\def\year{2018}\relax
\documentclass[letterpaper]{article} 
\usepackage{aaai18}  

\usepackage{multirow}
\usepackage{algorithm}
\usepackage{algorithmic}
\usepackage{amssymb}
\usepackage{amsmath}

\usepackage{times}  
\usepackage{helvet}  
\usepackage{courier}  
\usepackage{url}  
\usepackage{graphicx}  
\begin{document}
%
\title{Question Answering and Question Generation as Dual Tasks}
\author{AAAI Press\\
Association for the Advancement of Artificial Intelligence\\
2275 East Bayshore Road, Suite 160\\
Palo Alto, California 94303\\
}
\author{Duyu Tang$^\ddag$, Nan Duan$^\ddag$, Tao Qin$^\ddag$, Zhao Yan$^\dag$ \and Ming Zhou$^\ddag$ \\
	$^\ddag$Microsoft Research Asia, Beijing, China \\
	$^\dag$Beihang University, Beijing, China \\
	{\tt \{dutang,nanduan,taoqin,v-zhaoya,mingzhou\}@microsoft.com}}

\maketitle

\begin{abstract}
	We study the problem of joint question answering (QA) and question generation (QG) in this paper.
	Our intuition is that QA and QG have intrinsic connections and these two tasks could improve each other.
	On one side, the QA model judges whether the generated question of a QG model is relevant to the answer.
	On the other side, the QG model provides the probability of generating a question given the answer, which 
	is a useful evidence that 
	in turn facilitates QA.
	In this paper we regard QA and QG as dual tasks.
	We propose a training framework that trains the models of QA and QG simultaneously, and explicitly leverages their probabilistic correlation to guide the training process of both models.
	We implement a QG model based on sequence-to-sequence learning, and a QA model based on recurrent neural network.
	As all the components of the QA and QG models are differentiable, all the parameters involved in these two models could be conventionally  learned with back propagation.
	We conduct experiments on three datasets. Empirical results show that our training framework improves both QA and QG tasks.
	The improved QA model performs comparably with strong baseline approaches on all three datasets.
\end{abstract}

\section{Introduction}

Question answering (QA) and question generation (QG) are two fundamental tasks in natural language processing \cite{Manning1999,Jurafsky2000}.
Both tasks involve reasoning between a question sequence $q$ and an answer sentence $a$.
In this work, we take answer sentence selection \cite{yang2015wikiqa} as the QA task, which is a fundamental QA task and is very important for many applications such as search engine and conversational bots. 
The task of QA takes a question sentence $q$ and a list of candidate answer sentences as the input, and finds the top relevant answer sentence from the candidate list.
The task of QG takes a sentence $a$ as input, and generates a question sentence $q$ which could be answered by $a$. 

It is obvious that the input and the output of these two tasks are (almost) reverse, which is referred to as ``duality'' in this paper.
This duality connects QA and QG, and potentially could help these two tasks to improve each other.
%
Intuitively, QA could improve QG through measuring the relevance between the generated question and the answer.
This QA-specific signal could enhance the QG model to generate not only literally similar question string, but also the questions that could be answered by the answer.
In turn, QG could improve QA by providing additional signal which stands for the probability of generating a question given the answer.

Moreover, QA and QG have probabilistic correlation as both tasks relate to the joint probability between $q$ and $a$. 
Given a question-answer pair $\langle q, a \rangle$, the joint probability $P(q, a)$ can be computed in two equivalent ways.
\begin{equation}\label{equation:pqa}
P(q, a) = P(a) P(q|a) = P(q)P(a|q)
\end{equation}
The conditional distribution $P(q|a)$ is exactly the QG model, and the conditional distribution $P(a|q)$ is closely related to the QA model\footnote{In this work, our QA model is $f_{qa}(a,q;\theta_{qa})$. The conditional distribution $P(a|q)$ could be derived from the QA model, which will be detailed in the next section.}.
Existing studies typically learn the QA model and the QG model separately by minimizing their own loss functions, while ignoring the probabilistic correlation between them. 

Based on these considerations, we introduce a training framework that exploits the duality of QA and QG to improve both tasks.
There might be different ways of exploiting the duality of QA and QG. 
In this work, we leverage the probabilistic correlation between QA and QG as the regularization term to influence the training process of both tasks. 
Specifically, the training objective of our framework is to jointly learn the QA model parameterized by $\theta_{qa}$ and the QG model parameterized by $\theta_{qg}$ by minimizing their loss functions subject to the following constraint. 
\begin{equation}\label{equation:regular}
P_a(a) P(q|a;\theta_{qg}) = P_q(q)P(a|q;\theta_{qa})
\end{equation}
$P_a(a)$ and $P_q(q)$ are the language models for answer sentences and question sentences, respectively.

We examine the effectiveness of our training criterion by applying it to strong neural network based QA and QG models.
Specifically, we implement a generative QG model based on sequence-sequence learning, which takes an answer sentence as input and generates a question sentence in an end-to-end fashion.
We implement a discriminative QA model based on recurrent neural network, where both question and answer are represented as continuous vector in a sequential way.
As every component in the entire framework is differentiable, all the parameters could be conventionally  trained through back propagation.
We conduct experiments on three datasets \cite{yang2015wikiqa,rajpurkar-EtAl:2016:EMNLP2016,nguyen2016ms}. Empirical results show that our training framework improves both QA and QG tasks. The improved QA model performs comparably with strong baseline approaches on all three datasets.

\section{The Proposed Framework}
In this section, we first formulate the task of QA and QG, and then present the proposed algorithm for jointly training the QA and QG models.
We also describe 
the connections and differences between this work and existing studies.

\subsection{Task Definition and Notations}

This work involves two tasks, namely question answering (QA) and question generation (QG). 
There are different kinds of QA tasks in natural language processing community.
In this work, we take answer sentence selection \cite{yang2015wikiqa} as the QA task, which takes a question $q$ and a list of candidate answer sentences $A = \{a_1, a_2, ... , a_{|A|}\}$ as input, and outputs one answer sentence $a_i$ from the candidate list which has the largest probability to be the answer. This QA task is typically viewed as a ranking problem. Our QA model is abbreviated as $f_{qa}(a,q;\theta_{qa})$, which is parameterized by $\theta_{qa}$ and the output is a real-valued scalar.


The task of QG takes a sentence $a$ as input, and outputs a question $q$ which could be answered by $a$. 
In this work, we regard QG as a generation problem and develop a generative model based on sequence-to-sequence learning. Our QG model is abbreviated as $P_{qg}(q|a;\theta_{qg})$, which is parameterized by $\theta_{qg}$ and the output is the probability of generating a natural language question $q$.

\subsection{Algorithm Description}
We describe the proposed algorithm in this subsection.
Overall, the framework includes three components, namely a QA model, a QG model and a regularization term that reflects the duality of QA and QG. 
Accordingly, the training objective of our framework includes three parts, which is described in Algorithm 1. 

The QA specific objective aims to minimize the loss function $l_{qa}(f_{qa}(a,q;\theta_{qa}), label)$, 
where $label$ is 0 or 1 that indicates whether $a$ is the correct answer of $q$ or not.
Since the goal of a QA model is to predict whether a question-answer pair is correct or not, it is necessary to use negative QA pairs whose labels are zero.
The details about the QA model will be presented in the next section.

For each correct question-answer pair, the QG specific objective is to minimize the following loss function,
\begin{equation}
l_{qg}(q, a) = -log P_{qg}(q|a;\theta_{qg})
\end{equation}
where $a$ is the correct answer of $q$. 
The negative QA pairs are not necessary because the goal of a QG model is to generate the correct question for an answer.
The QG model will be described in the following section.

\begin{algorithm}[tb]
	\caption{Algorithm Description}
	\label{alg:example}
	\begin{algorithmic}
		\STATE {\bfseries Input:} Language models $P_a(a)$ and $P_q(q)$ for answer and question, respectively; hyper parameters $\lambda_q$ and $\lambda_a$; optimizer $opt$
		\STATE {\bfseries Output:} QA model $f_{qa}(a,q)$ parameterized by $\theta_{qa}$; QG model $P_{qg}(q|a)$ parameterized by $\theta_{qg}$
		\STATE
		\STATE Randomly initialize $\theta_{qa}$ and $\theta_{qg}$
		\REPEAT
		\STATE Get a minibatch of positive QA pairs $\langle q^p_i, a^p_i \rangle_{i=1}^m$, where $a_i$ is the answer of $q_i$;
		\STATE Get a minibatch of negative QA pairs $\langle q^n_i, a^n_i \rangle_{i=1}^m$, where $a^n_i$ is not the answer of $q^n_i$;
		
		\STATE Calculate the gradients for $\theta_{qa}$ and $\theta_{qg}$.
		\vspace{-0.3cm}
		\STATE \begin{align}\nonumber G_{qa} = \triangledown_{\theta_{qa}} &\frac{1}{m}\sum_{i = 1}^{m}[l_{qa}(f_{qa}(a^p_i,q^p_i;\theta_{qa}), 1) \\
		&\nonumber + l_{qa}(f_{qa}(a^n_i,q^n_i;\theta_{qa}),0) \\
		& +\lambda_al_{dual}(a^p_i,q^p_i;\theta_{qa}, \theta_{qg})]\end{align}
		\vspace{-0.8cm}
		\STATE \begin{align}\nonumber G_{qg} = \triangledown_{\theta_{qg}} &\frac{1}{m}\sum_{i = 1}^{m}[\ l_{qg}(q^p_i,a^p_i) \\& + \lambda_ql_{dual}(q^p_i,a^p_i;\theta_{qa}, \theta_{qg})]\end{align}
		\STATE Update $\theta_{qa}$ and $\theta_{qg}$
		\STATE $\theta_{qa} \leftarrow opt(\theta_{qa}, G_{qa})$, $\theta_{qg} \leftarrow opt(\theta_{qg}, G_{qg})$
		\UNTIL{models converged}
	\end{algorithmic}
\end{algorithm}

The third objective is the regularization term which satisfies the probabilistic duality constrains as given in Equation~\ref{equation:regular}.
Specifically, given a correct $\langle q, a \rangle$ pair, we would like to minimize the following loss function,
\begin{align} \nonumber
l_{dual}(a,q;\theta_{qa}, \theta_{qg}) &=  [logP_a(a) + log P(q|a;\theta_{qg})  \\
& - logP_q(q) - logP(a|q;\theta_{qa})]^2
\end{align}
where $P_a(a)$ and $P_q(q)$ are marginal distributions, which could be easily obtained through language model. $P(a|q;\theta_{qg})$ could also be easily calculated with the markov chain rule:
$P(q|a;\theta_{qg}) = \prod_{t=1}^{|q|} P(q_t|q_{<t}, a;\theta_{qg})$, where the function $P(q_t|q_{<t}, a;\theta_{qg})$ is the same with the decoder of the QG model (detailed in the following section).

However, the conditional probability $P(a|q;\theta_{qa})$ is different from the output of the QA model $f_{qa}(a,q;\theta_{qa})$. To address this, given a question $q$, we sample a set of answer sentences $A'$, and derive the conditional probability $P(a|q;\theta_{qa})$ based on our QA model with the following equation.
\begin{align}\nonumber
&P(a|q;\theta_{qa}) = \\
&\dfrac{exp(f_{qa}(a,q;\theta_{qa}))}{exp(f_{qa}(a,q;\theta_{qa})) + \sum_{a' \in A'} exp(f_{qa}(a',q;\theta_{qa}))}
\end{align}

In this way, we learn the models of QA and QG by minimizing the weighted combination between the original loss functions and the regularization term.



\subsection{Relationships with Existing Studies}
Our work differs from \cite{yang2017semi} in that they regard reading comprehension (RC) as the main task, and regard question generation as the auxiliary task to boost the main task RC. 
In our work, the roles of QA and QG are the same, and our algorithm enables QA and QG to improve the performance of each other simultaneously.
Our approach differs from Generative Domain-Adaptive Nets \cite{yang2017semi} in that we do not pretrain the QA model. Our QA and QG models are jointly learned from random initialization. 
Moreover, our QA task differs from RC in that the answer in our task is a sentence rather than a text span from a sentence.

Our approach is inspired by dual learning \cite{xia2016dual,xia2017dual}, which leverages the duality between two tasks to improve each other. Different from the dual learning \cite{xia2016dual} paradigm, our framework learns both models from scratch and does not need task-specific pretraining.
The recently introduced supervised dual learning \cite{xia2017dual} has been successfully applied to image recognition, machine translation and sentiment analysis. Our work could be viewed as the first work that leveraging the idea of supervised dual learning for question answering.
Our approach differs from Generative Adversarial Nets (GAN) \cite{goodfellow2014generative} in two respects. 
On one hand, the goal of original GAN is to learn a powerful generator, while the discriminative task is regarded as the auxiliary task. The roles of the two tasks in our framework are the same.
On the other hand, the discriminative task of GAN aims to distinguish between the real data and the artificially generated data, while we focus on the real QA task.


\section{The Question Answering Model}
We describe the details of the question answer (QA) model in this section. 
Overall, a QA model could be formulated as a function $f_{qa}(q, a;\theta_{qa})$ parameterized by $\theta_{qa}$ that maps a question-answer pair to a scalar. In the inference process, given a $q$ and a list of candidate answer sentences, $f_{qa}(q, a;\theta_{qa})$ is used to calculate the relevance between $q$ and every candidate $a$. 
The top ranked answer sentence is regarded as the output.

We develop a neural network based QA model.
Specifically, we first represent each word as a low dimensional and real-valued vector, also known as word embedding \cite{Bengio2003,Mikolov2013a,Pennington2014}.
Afterwards, we use recurrent neural network (RNN) to map a question of variable length to a fixed-length vector.
To avoid the problem of gradient vanishing, we use gated recurrent unit (GRU) \cite{cho-EtAl:2014:EMNLP2014} as the basic computation unit.
The approach recursively calculates the hidden vector $h_{t}$ based on the current word vector $e^q_t$ and the output vector $h_{t-1}$ in the last time step,
\begin{align}
&z_i = \sigma(W_{z}e^q_{i} + U_{z}{h}_{i-1})  \\
&r_i = \sigma(W_{r}e^q_{i} + U_{r}{h}_{i-1})  \\
&\widetilde{h}_i = \tanh(W_{h}e^q_{i} + U_{h}(r_i \odot {h}_{i-1}))  \\
&{h}_{i} = z_i \odot \widetilde{h}_i + (1-z_i) \odot {h}_{i-1}  
\end{align}
where $z_i$ and $r_i$ are update and reset gates of s, $\odot$ stands for element-wise multiplication, $\sigma$ is sigmoid function.
We use a bi-directional RNN to get the meaning of a question from both directions, and use the concatenation of two last hidden states as the final question vector $v_q$. We compute the answer sentence vector $v_a$ in the same way.

After obtaining $v_q$ and $v_a$, we implement a simple yet effective way to calculate the relevance between question-sentence pair.
Specifically, we represent a question-answer pair as the concatenation of four vectors, namely $v(q, a) =  [v_q; v_a; v_q \odot v_a ; e_{c(q,a)}]$, where $\odot$ means element-wise multiplication, $c(q,a)$ is the number of co-occurred words in $q$ and $a$. 
We observe that incorporating the embedding of the word co-occurrence $e^c_{c(q,a)}$ could empirically improve the QA performance. 
We use an additional embedding matrix $L_c \in \mathbb{R}^{d_c \times |V_c|}$, where $d_c$ is the dimension of word co-occurrence vector and $|V_c|$ is vocabulary size. 
The values of $L_c$ are jointly learned during training. The output scalar $f_{qa}(a,q)$ is calculated by feeding $v(q,a)$ to a linear layer followed by $tanh$. 
We feed $f_{qa}(a,q)$ to a $softmax$ layer and use negative log-likelihood as the QA specific loss function. The basic idea of this objective is to classify whether a given question-answer is correct or not.
We also implemented a ranking based loss function $max(0, 1 - f_{qa}(q,a) + f_{qa}(q,a^*))$, whose basic idea is to assign the correct QA pair a higher score than a randomly select QA pair. However, our empirical results showed that the ranking loss performed worse than the negative log-likelihood loss function. We use log-likelihood as the QA loss function in the experiment.

\begin{table*}[t]
	\centering
	\begin{tabular}{l|ccc|ccc|ccc}
		\hline
		\multirow{2}{*}{} & \multicolumn{3}{c|}{MARCO} & \multicolumn{3}{c|}{SQUAD} & \multicolumn{3}{c}{WikiQA}  \\
		\cline{2-10}
		& Train & Dev & Test & Train & Dev & Test& Train & Dev & Test \\
		\hline
		{\# questions}  & 82,326 & 4,806 & 5,241 & 87,341 & 5,273 & 5,279  & 1,040 & 140 & 293\\
		\# question-answer pairs & 676,193  & 39,510  & 42,850 & 440,573  & 26,442  & 26,604  & 20,360 & 2,733 & 6,165\\
		{Avg \# answers per question} & 8.21& 8.22&8.18 & 5.04 & 5.01 & 5.04  & 19.57 & 19.52 & 21.04\\
		{Avg length of questions} & 6.05& 6.05& 6.10 & {11.37} &11.57  &11.46 & 6.40& 6.46& 6.42\\
		{Avg length of answers} & 82.73& 82.54& 82.89 & {27.80} & 28.70  & 28.66 & 30.04& 29.65& 28.91\\
		\hline
	\end{tabular}
	\caption{Statistics of the MARCO, SQUAD and WikiQA datasets for answer sentence selection.}
	\label{table:statistic}
\end{table*}

\section{The Question Generation Model}
We describe the  question generation (QG) model in this section.
The model is inspired by the recent success of sequence-to-sequence learning in neural machine translation.
Specifically, the QG model first calculates the representation of the answer sentence with an encoder, and then takes the answer vector to generate a question in a sequential way with a decoder.
We will present the details of the encoder and the decoder, respectively.

The goal of the encoder is to represent a variable-length answer sentence ${a}$ as a fixed-length continuous vector.
The encoder could be implemented with different neural network architectures such as convolutional neural network \cite{kalchbrenner-blunsom:2013:EMNLP,meng2015encoding} and recurrent neural network (RNN) \cite{bahdanau2014neural,sutskever2014sequence}.
In this work, we use bidirectional RNN based on GRU unit, which is consistent with our QA model as described in Section 3.
The concatenation of the last hidden vectors from both directions is used as the output of the encoder, which is also used as the initial hidden state of the decoder.

The decoder takes the output of the encoder and generates the question sentence.
We implement a RNN based decoder, which works in a sequential way and generates one question word at each time step.
The decoder generates a word $q_{t}$ at each time step $t$ based on the representation of $a$ and the previously predicted question words $q_{<t}=\{q_1,q_2,...,q_{t-1}\}$.
This process is formulated as follows.

\begin{equation}
p(q|a)=\prod^{|q|}_{t=1}p(q_{t}|q_{<t},a)
\end{equation}

Specifically, we use an attention-based architecture \cite{luong-pham-manning:2015:EMNLP}, which selectively finds relevant information from the answer sentence when generating the question word. Therefore, the conditional probability is calculated as follows.
\begin{equation}
p(q_{t}|q_{<t},a)=f_{dec}(q_{t-1},s_{t}, c_t)
\end{equation}
where $s_{t}$ is the hidden state of GRU based RNN at time step $t$, and $c_t$ is the attention state at time step $t$.
The attention mechanism assigns a probability/weight to each hidden state in the encoder at one time step, and calculates the attention state $c_t$ through weighted averaging the hidden states of the encoder: $c_{t}=\sum^{|a|}_{i=1}\alpha_{\langle t,i\rangle}h_i$.
When calculating the attention weight of $h_i$ at time step $t$, we also take into account of the attention distribution in the last time step. Potentially, the model could remember which contexts from answer sentence have been used before, and does not repeatedly use these words to generate the question words.
\begin{align}
\alpha_{\langle t,i\rangle}=\frac{\exp{[z(s_{t},h_i,\sum^{N}_{j=1}\alpha_{\langle t-1,j\rangle}h_j)]}}{\sum^{H}_{i'=1}\exp{[z(s_{t},h_{i'},\sum^{N}_{j=1}\alpha_{\langle t-1,j\rangle}h_{j})]}} 
\end{align}

Afterwards, we feed the concatenation of $s_t$ and $c_t$ to a linear layer followed by a $softmax$ function.
The output dimension of the $softmax$ layer is equal to the number of top frequent question words (e.g. 30K or 50K) in the training data.
The output values of the $softmax$ layer form the probability distribution of the question words to be generated.
Furthermore, we observe that question sentences typically include informative but low-frequency words such as named entities or numbers.
These low-frequency words are closely related to the answer sentence but could not be well covered in the target vocabulary.
To address this, we add a simple yet effective post-processing step which replaces each ``unknown word'' with the most relevant word from the answer sentence. 
Following \cite{luong-EtAl:2015:ACL-IJCNLP}, we use the attention probability as the relevance score of each word from the answer sentence.
Copying mechanism \cite{gulcehre2016pointing,gu2016incorporating} is an alternative solution that adaptively determines whether the generated word comes from the target vocabulary or from the answer sentence.

Since every component of the QG model is differentiable, all the parameters could be learned in an end-to-end way with back propagation.
Given a question-answer pair $\langle q,a\rangle$, where $a$ is the correct answer of the question $q$, the training objective is to minimize the following negative log-likelihood.
\begin{equation}
l_{qg}(q,a)=-\sum^{|q|}_{t=1}\log[p(y_t|y_{<t},a)] 
\end{equation}
In the inference process, we use beam search to get the top-$K$ confident results, where $K$ is the beam size.
The inference process stops when the model generates the symbol $\langle eos \rangle$ which stands for the end of sentence.

\begin{table*}[t]
	\centering
	\begin{tabular}{l|ccc|ccc}
		\hline
		\multirow{2}{*}{Method} & \multicolumn{3}{c|}{MARCO} & \multicolumn{3}{c}{SQUAD} \\
		\cline{2-7}
		& MAP & MRR & P@1 & MAP & MRR & P@1 \\
		\hline
		WordCnt  & 0.3956 &0.4014&0.1789 & 0.8089&0.8168&0.6887\\
		WgtWordCnt  & 0.4223&	0.4287&0.2030 & 0.8714&0.8787&0.7958 \\
		CDSSM \cite{shen2014CDSSM}  & 0.4425 &0.4489 &0.2284 & 0.7978 & 0.8041 &0.6721 \\
		ABCNN \cite{yin2015abcnn} & 0.4691 & 0.4767 & 0.2629 & 0.8685 & 0.8750 & 0.7843 \\
		\hline
		Basic QA & 0.4712 & 0.4783 & 0.2628 & 0.8580 & 0.8647 & 0.7740 \\
		Dual QA & 0.4836 & 0.4911 & 0.2751 & 0.8643 & 0.8716 & 0.7814 \\
		\hline
	\end{tabular}
	\caption{QA Performance on the MARCO and SQUAD datasets.}
	\label{table:results-qa}
\end{table*}

\section{Experiment}
We describe the experimental setting and report empirical results in this section.

\subsection{Experimental Setting}


We conduct experiments on three datasets, including MARCO \cite{nguyen2016ms}, SQUAD \cite{rajpurkar-EtAl:2016:EMNLP2016}, and WikiQA \cite{yang2015wikiqa}.

The MARCO and SQUAD datasets
are originally developed for the reading comprehension (RC) task, the goal of which is to answer a question with a text span from a document. 
Despite our QA task (answer sentence selection) is different from RC, we use these two datasets because of two reasons. The first reason is that to our knowledge they are the QA datasets that contains largest manually labeled question-answer pairs. 
The second reason is that, we could derive two QA datasets for answer sentence selection from the original MARCO and SQUAD datasets, with an assumption that the answer sentences containing the correct answer span are correct, and vice versa.
We believe that our training framework could be easily applied to RC task, but we that is out of the focus of this work.

We also conduct experiments on WikiQA \cite{yang2015wikiqa}, which is a benchmark dataset for answer sentence selection. 
Despite its data size is relatively smaller compared with MARCO and SQUAD, we still apply our algorithm on this data and report empirical results to further compare with existing algorithms.

It is worth to note that a common characteristic of MARCO and SQUAD is that the ground truth of the test is invisible to the public. 
Therefore, we randomly split the original validation set into the dev set and the test set. 
The statistics of SQUAD and MARCO datasets are given in Table \ref{table:statistic}.
We use the official split of the WikiQA dataset.
 We apply exactly the same model to these three datasets. 


We evaluate our QA system with three standard evaluation metrics: \textit{Mean Average Precision (MAP)},  \textit{Mean Reciprocal Rank (MRR)} and \textit{Precision@1 (P@1)} \cite{manning2008ir}. 
It is hard to find a perfect way to automatically evaluate the performance of a QG system. In this work, we use BLEU-4~\cite{papineni2002bleu} score as the evaluation metric, which measures the overlap between the generated question and the ground truth.

%


\subsection{Implementation Details}

We train the parameters of the QA model and the QG model simultaneously. 
We randomly initialize the parameters in both models with a combination of the fan-in and fan-out \cite{glorot2010understanding}.
The parameters of word embedding matrices are shared in the QA model and the QG model.
In order to learn question and answer specific word meanings, we use two different embedding matrices for question words and answer words. The vocabularies are the most frequent 30K words from the questions and answers in the training data.
We set the  dimension of word embedding as 300, the hidden length of encoder and decoder in the QG model as 512, the hidden length of GRU in the QA model as 100, the dimension of word  co-occurrence embedding as 10, the vocabulary size of the word co-occurrence embedding as 10, the hidden length of the attention layer as 30. 
We initialize the learning rate as 2.0, and use AdaDelta~\cite{zeiler2012adadelta} to adaptively decrease the learning rate. 
We use mini-batch training, and empirically set the batch size as 64.
The sampled answer sentences do not come from the same passage. 
We get 10 batches (640 instances) and sort them by answer length for accelerating the training process. The negative samples come from these 640 instances, which are from different passages.

In this work, we use smoothed bigram language models as $p_a(a)$ and $p_q(q)$. We also tried trigram language model but did not get improved performance. 
Alternatively, one could also implement neural language model and jointly learn the parameters in the training process.

\subsection{Results and Analysis}
We first report results on the MARCO and SQUAD datasets. 
As the dataset is splitted by ourselves, we do not have previously reported results for comparison.
We compare with the following four baseline methods. 
It has been proven that word co-occurrence is a very simple yet effective feature for this task \cite{yang2015wikiqa,yin2015abcnn}, so the first two baselines are based on the word co-occurrence between a question sentence and the candidate answer sentence.
\textbf{WordCnt} and \textbf{WgtWordCnt} use unnormalized and normalized word co-occurrence.
The ranker in these two baselines are trained with with FastTree, which performs better than SVMRank and linear regression in our experiments.
We also compare with \textbf{CDSSM} \cite{shen2014CDSSM}, which is a very strong neural network approach to model the semantic relatedness of a sentence pair.
We further compare with \textbf{ABCNN} \cite{yin2015abcnn}, which has been proven very powerful in various sentence matching tasks.
\textbf{Basic QA} is our QA model which does not use the duality between QA and QG. 
Our ultimate model is abbreviated as \textbf{Dual QA}.

The QA performance on MARCO and SQUAD datasets are given in Table \ref{table:results-qa}.
We can find that CDSSM performs better than the word co-occurrence based method on MARCO dataset.
On the SQUAD dataset, Dual QA achieves the best performance among all these methods.
On the MARCO dataset, Dual QA performs comparably with ABCNN.

We can find that Dual QA still yields better accuracy than Basic QA, which shows the effectiveness of the joint training algorithm.
It is interesting that word co-occurrence based method (WgtWordCnt) is very strong and hard to beat on the MARCO dataset. Incorporating sophisticated features might obtain improved performance on both datasets, however, this is not the focus of this work and we leave it to future work.

\begin{table}[h]
	\centering
	\begin{tabular}{l|cc}
		\hline
		Method & MRR & MAP \\
		\hline
		CNN \cite{yang2015wikiqa} &0.6652& 0.6520 \\
		APCNN \cite{dos2016attentive} & 0.6957& 0.6886 \\
		NASM  \cite{miao2016neural} &0.7069 & 0.6886 \\
		ABCNN \cite{yin2015abcnn} & 0.7018 & 0.6921 \\
		\hline
		Basic QA & 0.6914 & 0.6747 \\
		Dual QA & 0.7002 & 0.6844 \\
		\hline
	\end{tabular}
	\caption{QA performance on the WikiQA dataset.}
	\label{table:results-qa-wikiqa}
\end{table}

\begin{table*}[t]\small
	\centering
	\begin{tabular}{p{3cm}|p{6.5cm}|p{3cm}|p{3cm}}
		\hline
		\textbf{question} & \textbf{correct answer} & \textbf{question generated by \ \ \ \ \ Dual QG} & \textbf{question generated by Basic QG}\\
		\hline
		\textit{what 's the name of the green space north of the center of newcastle ?} & \textit{Another green space in Newcastle is the Town Moor , lying immediately north of the city centre .} & \textit{what is the name of the green building in the city ?}  &\textit{ what is the name of the city of new haven ? }\\
		\hline
		\textit{for what purpose do organisms make peroxide and superoxide ?} & \textit{Parts of the immune system of higher organisms create peroxide , superoxide , and singlet oxygen to destroy invading microbes .} & \textit{what is the purpose of the immune system ?} & \textit{what is the main function of the immune system ?} \\
		\hline
		\textit{how much money was spent on other festivities in the bay area to help celebrate the coming super bowl 50 ?} & \textit{In addition , there are \$ 2 million worth of other ancillary events , including a week - long event at the Santa Clara Convention Center , a beer , wine and food festival at Bellomy Field at Santa Clara University , and a pep rally .}& \textit{how much of the beer is in the santa monica convention center ?} & \textit{what is the name of the beer in the santa monica center ?} \\
		\hline
	\end{tabular}
	\caption{Sampled examples from the SQUAD dataset.}
	\label{table:results-example}
\end{table*}

Results on the WikiQA dataset is given in Table \ref{table:results-qa-wikiqa}. 
On this dataset, previous studies typically report results based on their deep features plus the number of words that occur both in the question and in the answer \cite{yang2015wikiqa,yin2015abcnn}. We also follow this experimental protocol. We can find that our basic QA model is simple yet effective. The Dual QA model achieves comparably to strong baseline methods.

To give a quantitative evaluation of our training framework on the QG model, we report BLEU-4 scores on MARCO and SQUAD datasets.
The results of our QG model with or without using joint training are given in Table \ref{table:results-qg}.
We can find that, despite the overall BLEU-4 scores are relatively low, using our training algorithm could improve the performance of the QG model. 

\begin{table}[h]
	\centering
	\begin{tabular}{lccc}
		\hline
		{Method} & MARCO & SQUAD &WikiQA \\
		\hline
		Basic QG & 8.87 & 4.34 & 2.91\\
		Dual QG & 9.31 & 5.03 & 3.15\\
		\hline
	\end{tabular}
	\caption{QG performance (BLEU-4 scores) on  MARCO, SQUAD and WikiQA datasets.}
	\label{table:results-qg}
\end{table}

We would like to investigate how the joint training process improves the QA and QG models. To this end, we analyze the results of development set on the SQUAD dataset.
We randomly sample several cases that the Basic QA model gets the wrong answers while the Dual QA model obtains the correct results.
Examples are given in Table \ref{table:results-example}.
From these examples, we can find that the questions generated by Dual QG tend to have more word overlap with the correct question, despite sometimes the point of the question is not correct. 
For example, compared with the Basic QG model, the Dual QG model generates more informative words, such as ``green'' in the first example, ``purpose'' in the second example, and ``how much'' in the third example. 
We believe this helps QA because the QA model is trained to assign a higher score to the question which looks similar with the generated question.
It also helps QG because the QA model is trained to give a higher score to the real question-answer pair, so that generating more answer-alike words gives the generated question a higher QA score.

Despite the proposed training framework obtains some improvements on QA and QG, we believe the work could be further improved from several directions. 
We find that our QG model not always finds the point of the reference question. 
This is not surprising because the questions from these two reading comprehension datasets only focus on some spans of a sentence, rather than the entire sentence. Therefore, the source side (answer sentence) carries more information than the target side (question sentence). 
Moreover, we do not use the answer position information in our QG model.
Accordingly, the model may pay attention to the point which is different from the annotator's direction, and generates totally different questions.
We are aware of incorporating the position of the answer span could get improved performance \cite{zhou2017neural}, however, the focus of this work is a sentence level QA task rather than reading comprehension. 
Therefore, despite MARCO and SQUAD are of large scale, they are not the desirable datasets for investigating the duality of our QA and QG tasks.
Pushing forward this area also requires large scale sentence level QA datasets. 

\subsection{Discussion}
We would like to discuss our understanding about the duality of QA and QG, and also present our observations based on the experiments.

In this work, ``duality'' means that the QA task and the QG task are equally important. This characteristic makes our work different from Generative Domain-Adaptive Nets \cite{yang2017semi} and Generative Adversarial Nets (GAN) \cite{goodfellow2014generative}, both of which have a main task and regard another task as the auxiliary one.
There are different ways to leverage the ``duality'' of QA and QG to improve both tasks.
We categorize them into two groups. The first group is about the training process and the second group is about the inference process.
From this perspective, dual learning \cite{xia2016dual} is a solution that leverages the duality in the training process.
In particular, dual learning first pretrains the models for two tasks separately, and then iteratively fine-tunes the models.
Our work also belongs to the first group.
Our approach uses the duality as a regularization item to guide the learning of QA and QG models simultaneously from scratch.
After the QA and QG models are trained, we could also use the duality to improve the inference process, which falls into the second group.
The process could be conducted on separately trained models or the models that jointly trained with our approach.
This is reasonable because the QA model could directly add one feature to consider $q$ and $q'$, where $q'$ is the  question generated by the QG model. 
The first example in Table \ref{table:results-example} also motivates this direction.
Similarly, the QA model could give each $\langle q', a \rangle$ a score which could be assigned to each generated question $q'$.
In this work we do not apply the duality in the inference process. We leave it as a future plan. 

This work could be improved by refining every component involved in our framework.
For example, we use a simple yet effective QA model, which could be improved by using more complex neural network architectures \cite{hu2014convolutional,yin2015abcnn} or more external resources.
We use a smoothed language model for both question and answer sentences, which could be replaced by designed neural language models whose parameters are jointly learned together with the parameters in QA and QG models.
The QG model could be improved as well, for example, by developing more complex neural network architectures to take into account of more information about the answer sentence in the generation process.

In addition, it is also very important to investigate an automatic evaluation metric to effectively measure the performance of a QG system. 
BLEU score only measures the literal similarity between the generated question and the ground truth. 
However, it does not measure whether the question really looks like a question or not. 
A desirable evaluation system should also have the ability to judge whether the generated question could be answered by input sentence, even if the generated question use totally different words to express the meaning.

\section{Related Work}
Our work relates to existing studies on question answering (QA) and question generation (QG).

There are different types of QA tasks including text-level QA \cite{yu2014deep}, knowledge based QA \cite{berant2013semantic}, community based QA \cite{qiu2015convolutional} and the reading comprehension \cite{rajpurkar-EtAl:2016:EMNLP2016,nguyen2016ms}.
Our work belongs to text based QA where the answer is a sentence.
In recent years, neural network approaches \cite{hu2014convolutional,yu2014deep,yin2015abcnn} show promising ability in modeling the semantic relation between sentences and achieve strong performances on QA tasks. 

Question generation also draws a lot of attentions in recent years. 
QG is very necessary in real application as it is always time consuming to create large-scale QA datasets. 
In literature, \cite{yao2010question} use Minimal Recursion Semantics (MRS) to represent the meaning of a sentence, and then realize the MSR structure into a natural language question.
\cite{heilman2011automatic} present a overgenerate-and-rank framework consisting of three stages. They first transform a sentence into a simpler declarative statement, and then transform the statement to candidate questions by executing well-defined
syntactic transformations. Finally, a ranker is used to select the  questions of high-quality.
\cite{chali2015towards} focus on generating questions from a topic. 
They first get a list of texts related to the topic, and then generate questions by exploiting the named entity information and the predicate argument structures of the texts.
\cite{labutov2015deep} propose an ontology-crowd-relevance approach to generate questions from novel text. 
They encode the original text in a low-dimensional ontology, and then align the question templates obtained via crowd-sourcing to that space. A final ranker is used to select the top relevant templates.
There also exists some studies on generating questions from knowledge base \cite{song2016domain,serban-EtAl:2016:P16-1}. 
For example, \cite{serban-EtAl:2016:P16-1} develop a neural network approach which takes a knowledge fact (including a subject, an object, and a predicate) as input, and generates the question with a recurrent neural network.
Recent studies also investigate question generation for the reading comprehension task \cite{du2017question,zhou2017neural}. 
The approaches are typically based on the encoder-decoder framework, which could be conventionally learned in an end-to-end way.
As the answer is a text span from the sentence/passage, it is helpful to incorporate the position of the answer span \cite{zhou2017neural}.
In addition, the computer vision community also pays attention to generating natural language questions about an image \cite{mostafazadeh2016generating}.

\section{Conclusion}
We focus on jointly training the question answering (QA) model and the question generation (QG) model in this paper. 
We exploit the ``duality'' of QA and QG tasks, and introduce a training framework to leverage the probabilistic correlation between the two tasks.
In our approach, the ``duality'' is used as a regularization term to influence the learning of QA and QG models.
We implement simple yet effective QA and QG models, both of which are neural network based approaches.
Experimental results show that the proposed training framework improves both QA and QG on three datasets.

\subsubsection{Acknowledgments.}
We sincerely thank Wenpeng Yin for running the powerful ABCNN model on our setup. 
\bibliography{emnlp2017}

\begin{thebibliography}{}

\bibitem[\protect\citeauthoryear{Bahdanau, Cho, and
  Bengio}{2014}]{bahdanau2014neural}
Bahdanau, D.; Cho, K.; and Bengio, Y.
\newblock 2014.
\newblock Neural machine translation by jointly learning to align and
  translate.
\newblock {\em arXiv preprint arXiv:1409.0473}.

\bibitem[\protect\citeauthoryear{Bengio \bgroup et al\mbox.\egroup
  }{2003}]{Bengio2003}
Bengio, Y.; Ducharme, R.; Vincent, P.; and Janvin, C.
\newblock 2003.
\newblock A neural probabilistic language model.
\newblock {\em Journal of Machine Learning Research} 3:1137--1155.

\bibitem[\protect\citeauthoryear{Berant \bgroup et al\mbox.\egroup
  }{2013}]{berant2013semantic}
Berant, J.; Chou, A.; Frostig, R.; and Liang, P.
\newblock 2013.
\newblock Semantic parsing on freebase from question-answer pairs.
\newblock In {\em EMNLP}, volume~2, ~6.

\bibitem[\protect\citeauthoryear{Chali and Hasan}{2015}]{chali2015towards}
Chali, Y., and Hasan, S.~A.
\newblock 2015.
\newblock Towards topic-to-question generation.
\newblock {\em Computational Linguistics}.

\bibitem[\protect\citeauthoryear{Cho \bgroup et al\mbox.\egroup
  }{2014}]{cho-EtAl:2014:EMNLP2014}
Cho, K.; van Merrienboer, B.; Gulcehre, C.; Bahdanau, D.; Bougares, F.;
  Schwenk, H.; and Bengio, Y.
\newblock 2014.
\newblock Learning phrase representations using rnn encoder--decoder for
  statistical machine translation.
\newblock In {\em Proceedings of the 2014 Conference on Empirical Methods in
  Natural Language Processing (EMNLP)},  1724--1734.

\bibitem[\protect\citeauthoryear{dos Santos \bgroup et al\mbox.\egroup
  }{2016}]{dos2016attentive}
dos Santos, C.~N.; Tan, M.; Xiang, B.; and Zhou, B.
\newblock 2016.
\newblock Attentive pooling networks.
\newblock {\em CoRR, abs/1602.03609}.

\bibitem[\protect\citeauthoryear{Du, Shao, and Cardie}{2017}]{du2017question}
Du, X.; Shao, J.; and Cardie, C.
\newblock 2017.
\newblock Learning to ask: Neural question generation for reading
  comprehension.
\newblock In {\em Association for Computational Linguistics (ACL)}.

\bibitem[\protect\citeauthoryear{Glorot and
  Bengio}{2010}]{glorot2010understanding}
Glorot, X., and Bengio, Y.
\newblock 2010.
\newblock Understanding the difficulty of training deep feedforward neural
  networks.
\newblock In {\em Aistats}, volume~9,  249--256.

\bibitem[\protect\citeauthoryear{Goodfellow \bgroup et al\mbox.\egroup
  }{2014}]{goodfellow2014generative}
Goodfellow, I.; Pouget-Abadie, J.; Mirza, M.; Xu, B.; Warde-Farley, D.; Ozair,
  S.; Courville, A.; and Bengio, Y.
\newblock 2014.
\newblock Generative adversarial nets.
\newblock In {\em Advances in neural information processing systems},
  2672--2680.

\bibitem[\protect\citeauthoryear{Gu \bgroup et al\mbox.\egroup
  }{2016}]{gu2016incorporating}
Gu, J.; Lu, Z.; Li, H.; and Li, V.~O.
\newblock 2016.
\newblock Incorporating copying mechanism in sequence-to-sequence learning.
\newblock {\em arXiv preprint arXiv:1603.06393}.

\bibitem[\protect\citeauthoryear{Gulcehre \bgroup et al\mbox.\egroup
  }{2016}]{gulcehre2016pointing}
Gulcehre, C.; Ahn, S.; Nallapati, R.; Zhou, B.; and Bengio, Y.
\newblock 2016.
\newblock Pointing the unknown words.
\newblock {\em arXiv preprint arXiv:1603.08148}.

\bibitem[\protect\citeauthoryear{Heilman}{2011}]{heilman2011automatic}
Heilman, M.
\newblock 2011.
\newblock {\em Automatic factual question generation from text}.
\newblock Ph.D. Dissertation, Carnegie Mellon University.

\bibitem[\protect\citeauthoryear{Hu \bgroup et al\mbox.\egroup
  }{2014}]{hu2014convolutional}
Hu, B.; Lu, Z.; Li, H.; and Chen, Q.
\newblock 2014.
\newblock Convolutional neural network architectures for matching natural
  language sentences.
\newblock In {\em Advances in neural information processing systems},
  2042--2050.

\bibitem[\protect\citeauthoryear{Jurafsky and Martin}{2000}]{Jurafsky2000}
Jurafsky, D., and Martin, J.~H.
\newblock 2000.
\newblock Pearson education india.
\newblock {\em Speech and language processing}.

\bibitem[\protect\citeauthoryear{Kalchbrenner and
  Blunsom}{2013}]{kalchbrenner-blunsom:2013:EMNLP}
Kalchbrenner, N., and Blunsom, P.
\newblock 2013.
\newblock Recurrent continuous translation models.
\newblock In {\em Proceedings of the 2013 Conference on Empirical Methods in
  Natural Language Processing},  1700--1709.

\bibitem[\protect\citeauthoryear{Labutov, Basu, and
  Vanderwende}{2015}]{labutov2015deep}
Labutov, I.; Basu, S.; and Vanderwende, L.
\newblock 2015.
\newblock Deep questions without deep understanding.
\newblock In {\em ACL (1)},  889--898.

\bibitem[\protect\citeauthoryear{Luong \bgroup et al\mbox.\egroup
  }{2015}]{luong-EtAl:2015:ACL-IJCNLP}
Luong, T.; Sutskever, I.; Le, Q.; Vinyals, O.; and Zaremba, W.
\newblock 2015.
\newblock Addressing the rare word problem in neural machine translation.
\newblock In {\em Proceedings of the 53rd Annual Meeting of the Association for
  Computational Linguistics},  11--19.

\bibitem[\protect\citeauthoryear{Luong, Pham, and
  Manning}{2015}]{luong-pham-manning:2015:EMNLP}
Luong, T.; Pham, H.; and Manning, C.~D.
\newblock 2015.
\newblock Effective approaches to attention-based neural machine translation.
\newblock In {\em Proceedings of the 2015 Conference on Empirical Methods in
  Natural Language Processing},  1412--1421.

\bibitem[\protect\citeauthoryear{Manning and Sch{\"u}tze}{1999}]{Manning1999}
Manning, C.~D., and Sch{\"u}tze, H.
\newblock 1999.
\newblock Foundations of statistical natural language processing.
\newblock {\em MIT press}.

\bibitem[\protect\citeauthoryear{Manning \bgroup et al\mbox.\egroup
  }{2008}]{manning2008ir}
Manning, C.~D.; Raghavan, P.; Sch{\"u}tze, H.; et~al.
\newblock 2008.
\newblock {\em Introduction to information retrieval}, volume~1.
\newblock Cambridge university press Cambridge.

\bibitem[\protect\citeauthoryear{Meng \bgroup et al\mbox.\egroup
  }{2015}]{meng2015encoding}
Meng, F.; Lu, Z.; Wang, M.; Li, H.; Jiang, W.; and Liu, Q.
\newblock 2015.
\newblock Encoding source language with convolutional neural network for
  machine translation.
\newblock {\em arXiv preprint arXiv:1503.01838}.

\bibitem[\protect\citeauthoryear{Miao, Yu, and Blunsom}{2016}]{miao2016neural}
Miao, Y.; Yu, L.; and Blunsom, P.
\newblock 2016.
\newblock Neural variational inference for text processing.
\newblock In {\em International Conference on Machine Learning},  1727--1736.

\bibitem[\protect\citeauthoryear{Mikolov \bgroup et al\mbox.\egroup
  }{2013}]{Mikolov2013a}
Mikolov, T.; Sutskever, I.; Chen, K.; Corrado, G.~S.; and Dean, J.
\newblock 2013.
\newblock Distributed representations of words and phrases and their
  compositionality.
\newblock In {\em Advances in neural information processing systems},
  3111--3119.

\bibitem[\protect\citeauthoryear{Mostafazadeh \bgroup et al\mbox.\egroup
  }{2016}]{mostafazadeh2016generating}
Mostafazadeh, N.; Misra, I.; Devlin, J.; Mitchell, M.; He, X.; and Vanderwende,
  L.
\newblock 2016.
\newblock Generating natural questions about an image.
\newblock {\em arXiv preprint arXiv:1603.06059}.

\bibitem[\protect\citeauthoryear{Nguyen \bgroup et al\mbox.\egroup
  }{2016}]{nguyen2016ms}
Nguyen, T.; Rosenberg, M.; Song, X.; Gao, J.; Tiwary, S.; Majumder, R.; and
  Deng, L.
\newblock 2016.
\newblock Ms marco: A human generated machine reading comprehension dataset.
\newblock {\em arXiv preprint arXiv:1611.09268}.

\bibitem[\protect\citeauthoryear{Papineni \bgroup et al\mbox.\egroup
  }{2002}]{papineni2002bleu}
Papineni, K.; Roukos, S.; Ward, T.; and Zhu, W.-J.
\newblock 2002.
\newblock Bleu: a method for automatic evaluation of machine translation.
\newblock In {\em Proceedings of the 40th annual meeting on association for
  computational linguistics},  311--318.
\newblock Association for Computational Linguistics.

\bibitem[\protect\citeauthoryear{Pennington, Socher, and
  Manning}{2014}]{Pennington2014}
Pennington, J.; Socher, R.; and Manning, C.~D.
\newblock 2014.
\newblock Glove: Global vectors for word representation.
\newblock In {\em Proceedings of the 2014 Conference on Empirical Methods in
  Natural Language Processing (EMNLP)},  1532--1543.

\bibitem[\protect\citeauthoryear{Qiu and Huang}{2015}]{qiu2015convolutional}
Qiu, X., and Huang, X.
\newblock 2015.
\newblock Convolutional neural tensor network architecture for community-based
  question answering.
\newblock In {\em IJCAI},  1305--1311.

\bibitem[\protect\citeauthoryear{Rajpurkar \bgroup et al\mbox.\egroup
  }{2016}]{rajpurkar-EtAl:2016:EMNLP2016}
Rajpurkar, P.; Zhang, J.; Lopyrev, K.; and Liang, P.
\newblock 2016.
\newblock Squad: 100,000+ questions for machine comprehension of text.
\newblock In {\em Proceedings of the 2016 Conference on Empirical Methods in
  Natural Language Processing},  2383--2392.

\bibitem[\protect\citeauthoryear{Serban \bgroup et al\mbox.\egroup
  }{2016}]{serban-EtAl:2016:P16-1}
Serban, I.~V.; Garc\'{i}a-Dur\'{a}n, A.; Gulcehre, C.; Ahn, S.; Chandar, S.;
  Courville, A.; and Bengio, Y.
\newblock 2016.
\newblock Generating factoid questions with recurrent neural networks: The 30m
  factoid question-answer corpus.
\newblock In {\em Proceedings of the 54th Annual Meeting of the Association for
  Computational Linguistics},  588--598.

\bibitem[\protect\citeauthoryear{Shen \bgroup et al\mbox.\egroup
  }{2014}]{shen2014CDSSM}
Shen, Y.; He, X.; Gao, J.; Deng, L.; and Mesnil, G.
\newblock 2014.
\newblock A latent semantic model with convolutional-pooling structure for
  information retrieval.
\newblock In {\em Proceedings of the Conference on Information and Knowledge
  Management},  101--110.

\bibitem[\protect\citeauthoryear{Song and Zhao}{2016}]{song2016domain}
Song, L., and Zhao, L.
\newblock 2016.
\newblock Domain-specific question generation from a knowledge base.
\newblock {\em arXiv preprint arXiv:1610.03807}.

\bibitem[\protect\citeauthoryear{Sutskever, Vinyals, and
  Le}{2014}]{sutskever2014sequence}
Sutskever, I.; Vinyals, O.; and Le, Q.~V.
\newblock 2014.
\newblock Sequence to sequence learning with neural networks.
\newblock In {\em Advances in neural information processing systems},
  3104--3112.

\bibitem[\protect\citeauthoryear{Xia \bgroup et al\mbox.\egroup
  }{2016}]{xia2016dual}
Xia, Y.; He, D.; Qin, T.; Wang, L.; Yu, N.; Liu, T.-Y.; and Ma, W.-Y.
\newblock 2016.
\newblock Dual learning for machine translation.
\newblock {\em Advances in neural information processing systems}.

\bibitem[\protect\citeauthoryear{Xia \bgroup et al\mbox.\egroup
  }{2017}]{xia2017dual}
Xia, Y.; Qin, T.; Chen, W.; Bian, J.; Yu, N.; and Liu, T.-Y.
\newblock 2017.
\newblock Dual supervised learning.
\newblock {\em arXiv preprint arXiv:1707.00415}.

\bibitem[\protect\citeauthoryear{Yang \bgroup et al\mbox.\egroup
  }{2017}]{yang2017semi}
Yang, Z.; Hu, J.; Salakhutdinov, R.; and Cohen, W.~W.
\newblock 2017.
\newblock Semi-supervised qa with generative domain-adaptive nets.
\newblock {\em arXiv preprint arXiv:1702.02206}.

\bibitem[\protect\citeauthoryear{Yang, Yih, and Meek}{2015}]{yang2015wikiqa}
Yang, Y.; Yih, W.-t.; and Meek, C.
\newblock 2015.
\newblock Wikiqa: A challenge dataset for open-domain question answering.
\newblock In {\em EMNLP},  2013--2018.
\newblock Citeseer.

\bibitem[\protect\citeauthoryear{Yao}{2010}]{yao2010question}
Yao, X.
\newblock 2010.
\newblock Question generation with minimal recursion semantics.
\newblock Citeseer.

\bibitem[\protect\citeauthoryear{Yin \bgroup et al\mbox.\egroup
  }{2016}]{yin2015abcnn}
Yin, W.; Sch羹tze, H.; Xiang, B.; and Zhou, B.
\newblock 2016.
\newblock Abcnn: Attention-based convolutional neural network for modeling
  sentence pairs.
\newblock {\em Transactions of the Association for Computational Linguistics}
  4:259--272.

\bibitem[\protect\citeauthoryear{Yu \bgroup et al\mbox.\egroup
  }{2014}]{yu2014deep}
Yu, L.; Hermann, K.~M.; Blunsom, P.; and Pulman, S.
\newblock 2014.
\newblock Deep learning for answer sentence selection.
\newblock {\em arXiv preprint arXiv:1412.1632}.

\bibitem[\protect\citeauthoryear{Zeiler}{2012}]{zeiler2012adadelta}
Zeiler, M.~D.
\newblock 2012.
\newblock Adadelta: an adaptive learning rate method.
\newblock {\em arXiv preprint arXiv:1212.5701}.

\bibitem[\protect\citeauthoryear{Zhou \bgroup et al\mbox.\egroup
  }{2017}]{zhou2017neural}
Zhou, Q.; Yang, N.; Wei, F.; Tan, C.; Bao, H.; and Zhou, M.
\newblock 2017.
\newblock Neural question generation from text: A preliminary study.
\newblock {\em arXiv preprint arXiv:1704.01792}.

\end{thebibliography}
\bibliographystyle{aaai}

\end{document}